\documentclass[10pt,twocolumn,letterpaper]{article}

\usepackage[pagenumbers]{cvpr} % To force page numbers, e.g. for an arXiv version

\usepackage[accsupp]{axessibility}
\usepackage[dvipsnames]{xcolor}
\usepackage{multirow}
\usepackage{colortbl}
\definecolor{celadon}{rgb}{0.67, 0.88, 0.69}

\newcommand{\ourcolor}{\cellcolor{Cyan!15}}
\newcommand{\socolor}{\cellcolor{Gray!15}}
\newcommand{\tocolor}{\cellcolor{Gray!15}}
\usepackage{pifont}

\usepackage{fancyhdr}

\definecolor{cvprblue}{rgb}{0.21,0.49,0.74}
\usepackage[pagebackref,breaklinks,colorlinks,allcolors=cvprblue]{hyperref}
\usepackage{multirow}
\def\paperID{***} % *** Enter the Paper ID here
\def\confName{CVPR}
\def\confYear{2026}

\title{Learnable Motion-Focused Tokenization for Effective and Efficient Video Unsupervised Domain Adaptation}

\author{Tzu Ling Liu \qquad Ian Stavness \qquad Mrigank Rochan \\
	University of Saskatchewan, Canada
}

\begin{document}
\maketitle
\begin{abstract}
Video Unsupervised Domain Adaptation (VUDA) poses a significant challenge in action recognition, requiring the adaptation of a model from a labeled source domain to an unlabeled target domain. Despite recent advances, existing VUDA methods often fall short of fully supervised performance, a key reason being the prevalence of static and uninformative backgrounds that exacerbate domain shifts. Additionally, prior approaches largely overlook computational efficiency, limiting real-world adoption. To address these issues, we propose Learnable Motion-Focused Tokenization (LMFT) for VUDA. LMFT tokenizes video frames into patch tokens and learns to discard low-motion, redundant tokens, primarily corresponding to background regions, while retaining motion-rich, action-relevant tokens for adaptation. Extensive experiments on three standard VUDA benchmarks across 21 domain adaptation settings show that our VUDA framework with LMFT achieves state-of-the-art performance while significantly reducing computational overhead. LMFT thus enables VUDA that is both effective and computationally efficient.
\end{abstract}
    
\section{Introduction}\label{sec:intro}
Naively applying an action recognition model trained on labeled source domain videos to a novel unlabeled target domain often leads to severe performance degradation due to domain shifts arising from mismatches in both spatial appearance and complex temporal dynamics. To address this challenge, Video Unsupervised Domain Adaptation (VUDA) aims to leverage labeled source videos to align feature representations with those of unlabeled target videos. Earlier VUDA methods \cite{dann,mkmdd,da2022dual,sahoo2021contrast,chen2019temporal} mainly focus on temporal consistency techniques within ConvNet-based frameworks to mitigate such shifts. More recently, high-performing transformer architectures, particularly Vision Transformers (ViTs) \cite{dosovitskiy2020image}, have become the dominant choice for video understanding and have been adopted in leading VUDA methods \cite{zara2023unreasonable,unite,dacostaUnsupervisedDomainAdaptation2022}. While these ViT-based VUDA methods show strong promise and achieve state-of-the-art performance, they often fall short of fully supervised results on the target domain, exhibiting a substantial performance gap in many settings. Furthermore, computational efficiency, crucial for real-world adoption and deployment, remains largely neglected in current VUDA research.

While powerful, ViTs in VUDA process every spatio-temporal token extracted from both source and target domain videos, leading to two key challenges. First, not all tokens contribute meaningfully to action recognition; many correspond to static or background regions that are irrelevant to the action. For instance, the same action, such as running, may appear in different environments (e.g., indoor vs. outdoor), each with distinct background contexts. These variations amplify domain shifts and interfere with the transfer of motion-centric action semantics crucial for effective domain adaptation (DA). Second, because the computational cost of ViT self-attention scales quadratically with the number of tokens, processing all tokens, including those from redundant or uninformative regions, in both source and target videos is highly inefficient, hindering the practical adoption of VUDA in real-world applications.

To address these challenges, we propose \textit{Learnable Motion-Focused Tokenization} (LMFT) for more effective and efficient VUDA. LMFT explicitly identifies and retains regions within source and target video frames that exhibit meaningful, action-related motion dynamics, while discarding redundant low-motion or static background regions. Specifically, both source and target videos are initially divided into patch-level tokens, and pixel-wise motion differences are computed between temporally adjacent patches. Tokens with overall low motion, falling below a certain threshold (typically corresponding to static or redundant background content), are dropped, while high-motion, action-relevant tokens are retained and fed into the ViT-based DA framework. To automate and enable this token selection adaptive and content-dependent, we make the motion threshold \textit{learnable} using reinforcement learning (RL), since the token selection operation is inherently non-differentiable. This motion-focused token selection reduces background-induced domain shifts and enhances the model’s focus on transferable action-centric motion cues critical for \textit{effective} DA in action recognition. At the same time, it reduces the number of tokens fed into the ViT, making the adaptation process more computationally \textit{efficient} during both training and inference.

In summary, our contributions are threefold:
\textbf{(1)} We introduce Learnable Motion-Focused Tokenization (LMFT) for video DA, which selectively removes low-motion, redundant tokens while retaining motion-rich, action-relevant ones in both source and target videos. This targeted reduction of uninformative background tokens mitigates background-induced domain shifts, leading to more effective VUDA.
\textbf{(2)} We show that our VUDA framework with LMFT consistently outperforms the current state-of-the-art across three video DA benchmarks spanning 21 diverse adaptation settings.
\textbf{(3)} We present, to the best of our knowledge, the first exploration of computationally efficient VUDA, a critical yet previously overlooked dimension for real-world deployment. We provide a comprehensive efficiency analysis demonstrating that LMFT offers clear computational advantages over existing methods.

Together, these contributions highlight LMFT as a meaningful step toward effective and faster video DA suitable for practical applications.

\section{Related Work}\label{sec:related}
\noindent\textbf{VUDA for action recognition}. In recent years, VUDA has gained significant attention in action recognition. While applicable, image-based DA methods \cite{dann, mkmdd} offer limited improvement for videos, as they fail to model crucial temporal dynamics. This has motivated the development of video-specific adaptation frameworks. TA$^3$N \cite{chen2019temporal} employs a temporal relation module for dynamics and adversarial learning for spatial and temporal alignment. CoMix \cite{sahoo2021contrast} utilizes temporal contrastive learning to align cross-domain representations. UDAVT \cite{dacostaUnsupervisedDomainAdaptation2022} extends transformer architectures, including ViT, by applying the information bottleneck principle and cross-correlation matrix alignment. More recently, UNITE \cite{unite} employs masked modeling and self-training for cross-domain alignment within a ViT-based framework. However, these methods typically process full source and target domain video frames, which we argue is suboptimal for domain adaptation. Static or uninformative scene elements can interfere with the adaptation process. Moreover, the computational efficiency of VUDA remains largely unexplored. To address these challenges, we propose a method that retains motion-rich regions while discarding low-motion regions, thereby alleviating domain shift and enhancing computational efficiency for video DA.

\noindent\textbf{Token Reduction.}
Another line of research related to our work focuses on token reduction in transformer-based architectures, such as ViTs, to improve efficiency, as their computational cost grows significantly with the number of input tokens. For example, Token Merging (ToMe) \cite{bolya2022token} progressively fuses pairs of similar tokens based on their similarity scores for more efficient ViT processing. PruMerge \cite{prumerge} leverages internal ViT attention scores to guide pruning and merges discarded tokens into those retained ones, while DivPrune \cite{divprune} preserves representative tokens by maximizing token diversity using cosine distance. Another related approach is run-length tokenization (RLT) \cite{rlt}, which identifies and removes repeated runs of input patches over time by computing their differences. However, these token reduction methods are not specifically designed for video DA and typically rely on manually tuned pruning ratios or thresholding rules for token reduction, limiting their adaptability. In contrast, we introduce LMFT, a learnable token selection method that retains action-relevant tokens while discarding redundant tokens. LMFT not only enhances video DA but also improves computational efficiency through more optimized token reduction for scalable video DA.

\vspace{-2pt}
\section{Preliminaries}\label{sec:preliminaries}
\vspace{-5pt}
\begin{figure*}[t]
    \centering
    \includegraphics[width=\textwidth]{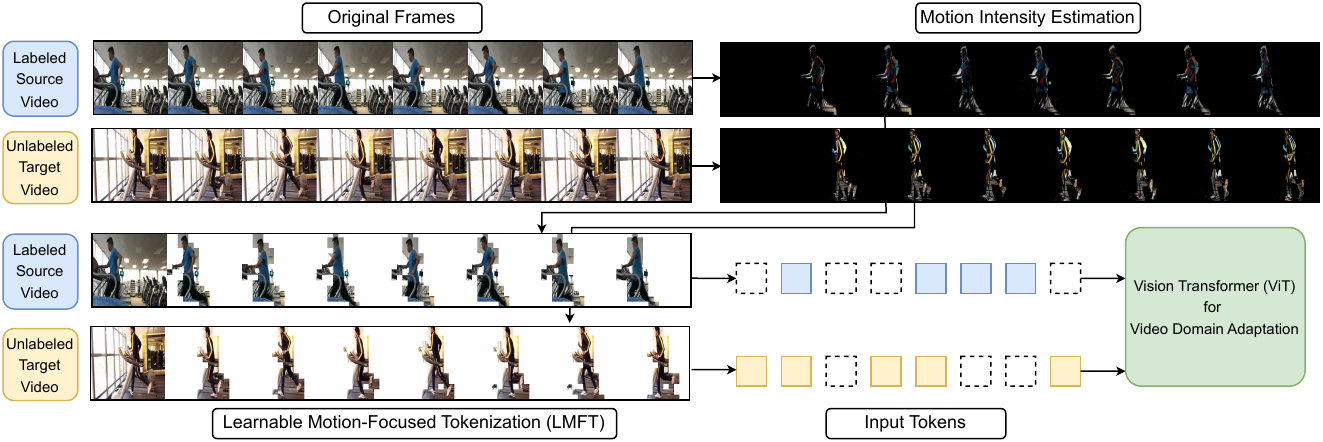}
    %\vspace{-15pt}
    \caption{
    Overview of LMFT. For both source and target videos, LMFT tokenizes frames into patch tokens, computes the L1 distance between consecutive temporal tokens, and discards those with differences below a learnable threshold $\tau$, which primarily correspond to redundant and static background regions. The remaining motion-rich action-relevant tokens are then fed into the ViT-based DA framework.
    }
    \vspace{-10pt}
    \label{figure:mft}
\end{figure*}

\noindent\textbf{Problem.} 
We address the problem of VUDA for action recognition.
Formally, we are given a labeled source-domain video dataset 
$\mathcal{V}_S = \{\mathbf{v}_i^S\}_{i=1}^{m}$ with corresponding action labels 
$\mathcal{A}_S = \{a_i^S\}_{i=1}^{m}$, and an unlabeled target-domain dataset 
$\mathcal{V}_T = \{\mathbf{v}_i^T\}_{i=1}^{n}$, where $m$ and $n$ denote the number of videos in the source and target domains, respectively. 
Both domains share an identical set of $C$ action categories, i.e., $\{1, 2, \ldots, C\}$.
The primary challenge arises from the domain discrepancy between $\mathcal{V}_S$ and $\mathcal{V}_T$, which typically causes models trained solely on the source domain to perform poorly on the target domain. 
We aim to train a ViT-based network $f_{\phi}$ that utilizes labeled source data and unlabeled target data for effective and efficient DA, minimizing classification errors on the target domain.

\noindent\textbf{Video Tokenization in ViT.}
Recently, ViTs \cite{dosovitskiy2020image} have been increasingly applied to video understanding tasks, including action recognition \cite{xing2023svformer, khan2024drone, huang2025sefar}. A fundamental prerequisite for applying ViT to video data is tokenization, which converts spatiotemporal patches into a sequence of embeddings suitable for ViT processing. 
Specifically, let $\mathbf{V} \in \mathbb{R}^{T \times C \times H \times W}$ denote an input video, where $T$, $C$, $H$, and $W$ correspond to the temporal, channel, height, and width dimensions, respectively.  
The standard tokenization scheme for video ViTs first partitions $\mathbf{V}$ into a 3D grid of non-overlapping spatiotemporal patches. Each patch, denoted as $\mathbf{P} \in \mathbb{R}^{t_p \times C \times p \times p}$, spans $t_p$ frames and has a spatial resolution of $p \times p$. 
We index each patch by its temporal position $t$ and its spatial grid indices $(x, y)$, denoted as $\mathbf{P}_t^{x,y}$.
Each patch is then treated as a token corresponding to a distinct spatiotemporal location in the video.
This partitioning transforms $\mathbf{V}$ into a collection of spatiotemporal patches, $\mathbf{V} \mapsto \{\mathbf{P}_t^{x,y}\}$, yielding a total of $N = N_t \times N_x \times N_y$ patches, where $N_t = T / t_p$, $N_x = H / p$, and $N_y = W / p$ denote the patch counts along the temporal, height, and width dimensions, respectively. This standard scheme results in the same number of tokens per video, regardless of content.

This content-agnostic design poses a major limitation: it produces a fixed set of $N$ tokens irrespective of the video’s spatiotemporal dynamics. Recent video DA methods \cite{zara2023unreasonable, unite, dacostaUnsupervisedDomainAdaptation2022} adopt this standard scheme, but we argue it is suboptimal for two reasons. First, many tokens correspond to static or low-motion backgrounds that are irrelevant to the action, obscuring the transferable, action-centric semantics essential for effective DA.
Second, it is computationally inefficient, as the ViT must process all tokens regardless of their relevance.
To address these issues, we propose LMFT for video DA. LMFT dynamically removes tokens associated with low-motion, redundant regions while retaining motion-rich, action-relevant ones, leading to more effective and efficient DA for action recognition.

\vspace{-5pt}
\section{Approach}\label{sec:approach}
\vspace{-5pt}
We propose a joint training framework, utilizing both labeled source data and unlabeled target data, to adapt the model $f_{\phi}$ to target domain.
Our framework has two key components: (1) LMFT, a module that dynamically selects motion-rich tokens and discards low-motion ones of each input video from source and target domains, (2) VUDA with LMFT, a ViT-based network that utilizes the selected tokens from LMFT for video DA and simultaneously optimizes the LMFT module.

\vspace{-5pt}
\subsection{Learnable Motion-Focused Tokenization (LMFT)}
\label{sec:lmft}
In LMFT (Fig.~\ref{figure:mft}), we first tokenize each input video from the source and target domains into patch-level tokens. We then compute the temporal motion to quantify the motion intensity of each token. Based on this estimation, LMFT employs a learnable token selection criterion using RL. Next, low-motion tokens are discarded, and only motion-rich tokens are fed into our ViT-based network for DA.

\noindent\textbf{Motion Intensity Estimation.}
To identify motion-rich tokens of a video, we first measure the temporal variation at each spatial location \cite{rlt}. Recall that $\mathbf{P}_t^{x,y} \in \mathbb{R}^{t_p \times C \times p \times p}$ denotes the $t$-th spatiotemporal patch at location $(x,y)$.
First, we compute a temporally-averaged representative patch, $\bar{\mathbf{P}}_t^{x,y} \in \mathbb{R}^{C \times p \times p}$, by averaging the original patch $\mathbf{P}_t^{x,y}$ along its temporal dimension.
Subsequently, for each spatial location $(x,y)$, we compute the pixel-wise motion difference $\mathbf{D}_t^{x,y}$ between \emph{adjacent} representative patches in time using the $L_1$ distance:
\begin{equation}
\mathbf{D}_t^{x,y} = \left| \bar{\mathbf{P}}_{t+1}^{x,y} - \bar{\mathbf{P}}_t^{x,y} \right|, \quad \text{for } t \in \{1, \dots, N_t-1\}
\end{equation}
This yields $N_t-1$ difference maps, where $\{\mathbf{D}_t^{x,y}\} \in \mathbb{R}^{C \times p \times p}$, representing the changes between consecutive patches.
Next, we define a scalar patch motion energy, $\textbf{E}_t^{x,y}$, by applying a global average pooling over the channel and spatial dimensions of each difference map $\mathbf{D}_t^{x,y}$ as:
% \begin{equation}
$
\textbf{E}_t^{x,y} = \frac{1}{C \cdot p \cdot p} \sum_{c, i, j} \mathbf{D}_t^{x,y}[c, i, j].
$
% \end{equation}
This $\textbf{E}_t^{x,y}$ quantifies the motion intensity at location $(x,y)$ between time intervals $t$ and $t+1$.
To make motion magnitudes comparable across videos, we normalize the energy values within each video using min–max normalization, yielding $\tilde{\textbf{E}}_t^{x,y} \in [0,1]$.
The sequence of normalized energies across time is denoted as $\tilde{\mathbf{E}}^{x,y} = [\tilde{\textbf{E}}_1^{x,y}, \ldots, \tilde{\textbf{E}}_{N_t-1}^{x,y}]$.
Since the first temporal segment lacks a preceding patch for differencing, we prepend a constant value of one to retain its information, forming the full motion energy sequence as
$\mathbf{E}_{\text{all}}^{x,y} = \text{concat}([\mathbf{1}, \tilde{\mathbf{E}}^{x,y}])$. 
This ensures the first temporal segment is always retained in subsequent motion-based token selection.

\noindent\textbf{Motion-Focused Token Selection.}
Note that for each spatial location $(x,y)$, we define the motion energy sequence
$\mathbf{E}_{\text{all}}^{x,y}$, 
which has a temporal length of $N_t$.
We then aggregate all $N_x \times N_y$ spatial sequences into a unified motion energy tensor, 
denoted as $\mathbf{E}_{\text{motion}} \in \mathbb{R}^{N_t \times N_x \times N_y}$.
Each element $(\mathbf{E}_{\text{motion}})_{t,x,y}$ represents the motion energy associated with the token 
at temporal segment $t$ and spatial location $(x,y)$.
Given this motion energy tensor $\mathbf{E}_{\text{motion}}$, we derive a binary selection mask 
$\mathbf{M} \in \{0,1\}^{N_t \times N_x \times N_y}$ to identify motion-rich tokens of the video, where each element $\mathbf{M}_{t,x,y}$ indicates whether the token at spatiotemporal position $(t,x,y)$ is retained.
A token is retained if its motion energy exceeds the motion-focused threshold $\tau \in (0,1)$, and is dropped otherwise:
\begin{equation}
\mathbf{M}_{t,x,y} =
\begin{cases}
1, & \text{if } (\mathbf{E}_{\text{motion}})_{t,x,y} > \tau, \\
0, & \text{otherwise.}
\end{cases}
\label{eq:selection_mask}
\end{equation}
With this mask, tokens $\mathbf{P}_t^{x,y}$ corresponding to $\mathbf{M}_{t,x,y}=1$ are identified as motion-rich and retained for subsequent processing, whereas those with $\mathbf{M}_{t,x,y}=0$ are dropped.
This threshold-based rule provides a transparent and interpretable mechanism:
a larger threshold $\tau$ enforces more aggressive dropping, whereas a smaller $\tau$ yields a denser set of retained tokens.
Next, we extend this formulation by making $\tau$ a learnable variable, allowing the model to optimally adjust to diverse motion patterns and domains.

\noindent\textbf{Learnable Motion-Focused Threshold $\tau$.} 
While the threshold $\tau$ can be manually tuned, this approach is labor-intensive and may not generalize across diverse action motion patterns and domains. By making it learnable, the network can optimize the threshold in an end-to-end manner, enabling adaptive, data-driven token selection for improved DA.
However, directly learning $\tau$ via standard backpropagation is non-trivial, since the decision to retain or drop tokens is based on a hard comparison, which is inherently non-differentiable. 
Consequently, we employ policy-gradient reinforcement learning (RL) \cite{williams1992simple, 2025bypass} for learning optimal $\tau$.
Our goal is to learn a stochastic policy $\pi_\theta(\tau)$ that samples an optimal threshold $\tau$. To ensure $\tau$ is bounded within $(0,1)$, we implement this policy as a logistic-normal distribution, parameterized by $\theta = \{\mu, \log\sigma\}$.
We parameterize the standard deviation using its logarithm, $\log\sigma$, which allows for unconstrained optimization in $\mathbb{R}$ while ensuring the standard deviation itself remains positive via $\sigma = \exp(\log\sigma)$.
The policy is trained to maximize a reward signal $R(\tau)$, which is designed to balance adaptation accuracy with computational efficiency (formally defined in Sec.~\ref{sec:vuda}). We formulate this objective as maximizing the expected reward:
% \begin{equation}
$
\mathcal{J}(\theta) = \mathbb{E}{\tau \sim \pi_\theta}[R(\tau)].
$
% \end{equation}
Given the non-differentiable nature of $R(\tau)$, we optimize policy parameters $\theta$ using REINFORCE gradient estimator:
\begin{equation}
\nabla_\theta \mathcal{J}(\theta)= \mathbb{E}{\tau \sim \pi_\theta}\big[(R(\tau) - b)\nabla_\theta \log \pi_\theta(\tau)\big],
\label{eq:policy_gradient}
\end{equation} 
where $b$ is an exponentially moving average ($b \leftarrow 0.9b + 0.1R(\tau)$)  baseline employed to reduce gradient variance.
To perform the REINFORCE update, we first sample an action $\tau \sim \pi_\theta$. This is achieved by sampling a latent variable $u \sim \mathcal{N}(\mu, \sigma^2)$ and applying the sigmoid transformation $\tau = \mathrm{sigmoid}(u)$. 
Next, we compute the gradient of the log-policy, $\nabla_\theta \log \pi_\theta(\tau)$. To obtain the analytical form of $\log \pi_\theta(\tau)$, we apply the change of variables formula based on this transformation: 
\begin{equation}
\log \pi_\theta(\tau) = \log \mathcal{N}\big(\text{logit}(\tau); \mu, \sigma^2\big) - \log(\tau) - \log(1-\tau)
\end{equation}
where $\text{logit}(\tau) = \log(\tau / (1-\tau))$ and $\mathcal{N}(\cdot; \mu, \sigma^2)$ is the probability density function of the Gaussian distribution. This provides the necessary component for the gradient estimator in Eq.~\ref{eq:policy_gradient}.
The overhead of this procedure is negligible, as only a single scalar $\tau$ is sampled and two parameters are updated via a lightweight REINFORCE step.
Through this process, the policy learns to infer an optimal motion-focused threshold $\tau$ (Eq. \ref{eq:selection_mask}) that adaptively regulates token selection based on the reward signal.
Next, the tokens selected by LMFT are fed into the ViT network for VUDA.

\subsection{VUDA with LMFT}\label{sec:vuda} 
Our VUDA with LMFT has the following key aspects:

\noindent\textbf{Pseudo-Label Generation.}
We employ a pretrained vision-language model, CLIP (ViT-B/16)~\cite{radford2021learning}, to generate pseudo-labels in a zero-shot fashion.
First, we construct a set of class-specific text embeddings, $\mathbf{L} = \{\mathbf{l}_c\}_{c=1}^C$, by feeding text prompts of the template “a video of a person \{action class\}” into the text encoder for all $C$ categories. For each target video $\mathbf{v}_i^T$, we derive a video-level representation $\mathbf{z}_i^T$ by passing its frames through the vision encoder and applying temporal average pooling. Subsequently, we compute the cosine similarity between $\mathbf{z}_i^T$ and 
the text embedding set $\mathbf{L}$. We apply a softmax function to the similarity scores to obtain the class probability distribution $\mathbf{q}_i^T$.

\noindent\textbf{Confidence-Based Filtering.}
To mitigate the impact of noisy pseudo-labels, we employ a confidence-based filtering strategy to curate a high-quality pseudo-labeled dataset. A target sample $\mathbf{v}_i^T$ is retained only if its maximum class probability exceeds a predefined confidence threshold $\gamma_c$, i.e., $\max(\mathbf{q}_i^T) > \gamma_c$. For such high-confidence samples, the class index corresponding to this maximum probability is assigned as the pseudo-label ${a'}^T_i$. This process yields a filtered target subset $\mathcal{V}'_T \subset \mathcal{V}_T$ and its corresponding pseudo-label set $\mathcal{A}'_T = \{ {a'}^T_i \}_{i=1}^{n'}$, which are utilized for subsequent training.
Our ViT model, $f_{\phi}$, is trained using labeled source videos and pseudo-labeled target videos.

\noindent\textbf{Target Adaptation.}
To adapt the model to the target domain, we define our joint training objective as a combination of source-domain supervision and target-domain pseudo-label supervision. In each training iteration, every video sampled from both the source domain ($\mathcal{V}_S$) and the filtered target domain ($\mathcal{V}'_T$) is first processed by LMFT. The selected motion-rich tokens are then forwarded to $f_{\phi}$, a ViT-B/16 network, to obtain predictions and compute the losses.
Specifically, the source domain loss $\mathcal{L}_{\text{s}}$ is the standard cross-entropy loss computed on $(\mathcal{V}_S, \mathcal{A}_S)$, while the target domain loss $\mathcal{L}{\text{t}}$ applies the same criterion on $(\mathcal{V}'_T, \mathcal{A}'_T)$:
\begin{equation}
\begin{split}
\mathcal{L}_{\text{s}} &=
\mathbb{E}_{(\mathbf{v}^S, a^S) \sim (\mathcal{V}_S, \mathcal{A}_S)}
\left[\mathcal{L}_{\text{CE}}\big(f_{\phi}(\mathbf{v}^S), a^S\big)\right], \\[4pt]
\mathcal{L}_{\text{t}} &=
\mathbb{E}_{(\mathbf{v}^T, a'^T) \sim (\mathcal{V}'_T, \mathcal{A}'_T)}
\left[\mathcal{L}_{\text{CE}}\big(f_{\phi}(\mathbf{v}^T), {a'}^T\big)\right]. \\[4pt]
\end{split}
\label{eqn:st_loss}
\end{equation}
The overall DA loss $\mathcal{L}_{\text{da}}$ combines both losses as:
\begin{equation}
\begin{split}
\mathcal{L}_{\text{da}} &= \mathcal{L}_{\text{s}} + \lambda_t \, \mathcal{L}_{\text{t}},
\end{split}
\label{eqn:primary_loss}
\end{equation}
where $\lambda_t > 0$ regulates their contributions.

\noindent\textbf{RL Reward Integration.}
During training, we define the policy $\pi_\theta$ that is optimized via RL (see Sec.~\ref{sec:lmft}) to dynamically adjust the motion-focused threshold $\tau$ (in Eq.\ref{eq:selection_mask}) for an optimal trade-off between DA performance and computational efficiency. Specifically, we define distinct reward components for the source and target domains to reflect both DA performance and efficiency:

\begin{equation}
R_{\text{src}} = -\lambda_{\mathcal{L}}\mathcal{L}_s - (1 - \rho_s),\quad R_{\text{tgt}} = -\lambda_{\mathcal{L}}\mathcal{L}_t - (1 - \rho_t),
\label{eq:reward}
\end{equation}
where $\mathcal{L}_{\text{s}}$ and $\mathcal{L}_{\text{t}}$ denote the source domain and target domain losses, $\rho_{s}$, $\rho_{t}$ are the corresponding token dropping ratios, defined as the proportion of tokens dropped relative to the total number of tokens in the video, and $\lambda_{\mathcal{L}}$ is a balancing hyperparameter. The total reward used for the policy update is $R(\tau) = R_{\text{src}} + R_{\text{tgt}}$.

\noindent\textbf{Testing.}
At test time, we adopt a deterministic motion-focused threshold $\hat{\tau}$ in LMFT for stable predictions. Since the sigmoid function is nonlinear, applying it to the mean $\mu$ alone (i.e., $\mathrm{sigmoid}(\mu)$) produces a biased estimate of the policy's expected value. To faithfully represent the stochastic policy (Eq.~\ref{eq:policy_gradient}) optimized during training, we set $\hat{\tau}$ to the expected value of the action distribution:
\begin{equation}
\hat{\tau} = \mathbb{E}_{\tau \sim \pi_\theta}[\tau] = \mathbb{E}_{\epsilon \sim \mathcal{N}(0,1)} \big[\mathrm{sigmoid}(\mu + \sigma \epsilon)\big].
\label{eqn:inference_exp}
\end{equation}
Since this expectation lacks a closed-form solution, we obtain it via Monte Carlo estimation. 
Specifically, we draw $K$ samples $\{\epsilon_k\}_{k=1}^K$ from $\mathcal{N}(0,1)$ and compute the deterministic threshold as:
$
% \begin{equation}
\hat{\tau} = \frac{1}{K} \sum_{k=1}^K \mathrm{sigmoid}(\mu + \sigma \epsilon_k).
$
This $\hat{\tau}$ is precomputed once after training, which simulates the learned policy distribution without incurring any deployment overhead. In our experiments, we set $K=100$.
Then, during the testing, each target video is processed by LMFT, which selects motion-rich, action-relevant tokens based on the threshold $\hat{\tau}$. The resulting subset of tokens is then fed into the trained ViT model $f_{\phi}$ to produce the final action prediction.

\section{Experiments}\label{sec:exp}

\begin{table*}[!ht]
\begin{center}
\footnotesize
\resizebox{1.0\linewidth}{!}{
\begin{tabular}{clccc|ccc|ccc|ccc|c}
\toprule
& \multirow{2}{*}{\textbf{Method}} & \multicolumn{12}{c}{\textbf{Accuracy (Top-1\%)}} \\
& & H→A & M→A & K→A & A→H & M→H & K→H & H→M & A→M & K→M & M→K & H→K & A→K &  \textbf{Avg.} \\
\midrule
& \socolor Source Only & \socolor 45.7 & \socolor \textbf{55.2} & \socolor 38.1 & \socolor 51.7 & \socolor 68.8 & \socolor 47.1 & \socolor 47.5 & \socolor 46.5 & \socolor 37.0 & \socolor 79.4 & \socolor 57.0 & \socolor 53.9 & \socolor 52.3 \\
\midrule    
\scriptsize \parbox[t]{2mm}{\multirow{1}{*}{\rotatebox{90}{ZS}}}
& CLIP (ViT-B/16) \cite{radford2021learning} & 36.5 & 36.5 & 36.5 & 60.0 & 60.0 & 60.0 & 48.5 & 48.5 & 48.5 & 68.1 & 68.1 & 68.1 & 53.3 \\
\midrule
\scriptsize \parbox[t]{2mm}{\multirow{3}{*}{\rotatebox{90}{SFVUDA}}}
& ATCoN \cite{atcon} & 17.9 & 27.2 & 17.2 & 26.7 & 47.3 & 48.2 & 30.7 & 17.2 & 32.5 & 57.7 & 48.5 & 31.0 & 33.5 \\
& EXTERN \cite{extern} & 26.2 & 18.1 & 23.9 & 26.2 & 53.7 & 55.8 & 40.7 & 18.2 & 35.2 & 68.1 & 57.6 & 51.4 & 39.6 \\
& STHC \cite{li2023source} & 15.5 & 48.7 & 34.8 & 18.4 & 56.3 & 76.6 & 13.8 & 39.8 & 50.1 & 44.6 & 27.3 & 44.7 & 39.2 \\
& DALL-V \cite{zara2023unreasonable} & 24.0 & 24.0 & 24.0 & 57.9 & 65.4 & 52.5 & 47.0 & 45.7 & 47.0 & 78.1 & 76.7 & 75.0 & 51.4 \\
\midrule
\scriptsize \parbox[t]{2mm}{\multirow{5}{*}{\rotatebox{90}{VUDA}}} & DANN \cite{dann} & 14.2 & 22.8 & 21.2 & 20.1 & 43.3 & 37.5 & 29.5 & 19.7 & 21.7 & 58.8 & 38.2 & 27.0 & 29.5 \\
& MK-MMD \cite{mkmdd} & 20.3 & 21.0 & 21.7 & 18.7 & 50.4 & 36.2 & 25.7 & 18.0 & 24.0 & 58.5 & 33.8 & 26.1 & 29.5 \\
& TA$^3$N \cite{chen2019temporal} & 14.4 & 21.6 & 19.9 & 14.9 & 43.0 & 37.7 & 25.7 & 15.6 & 31.5 & 55.5 & 38.4 & 23.4 & 28.5 \\
&  UNITE \cite{unite} &  48.0 &  44.1 &  37.5 &  67.9 &  \textbf{74.2} &  \textbf{65.8} & 51.8 &  50.0 &  48.0 &  \textbf{89.9} &  69.9 &  63.6 &  59.2 \\
& \ourcolor Ours w/o LMFT  & \ourcolor 43.6 & \ourcolor 45.9 & \ourcolor 39.4 & \ourcolor 68.3 & \ourcolor 72.1 & \ourcolor 57.5 & \ourcolor 57.5 & \ourcolor \textbf{54.8} & \ourcolor 53.5 & \ourcolor 84.0 & \ourcolor 81.8 & \ourcolor 85.4 & \ourcolor 62.0 \\
& \ourcolor Ours & \ourcolor \textbf{47.8} & \ourcolor 46.6 & \ourcolor \textbf{44.9} & \ourcolor \textbf{74.6} & \ourcolor \textbf{74.2} & \ourcolor 59.2 & \ourcolor \textbf{60.0} & \ourcolor 48.3 & \ourcolor \textbf{54.8} & \ourcolor 83.7 & \ourcolor \textbf{83.6} & \ourcolor \textbf{86.6} & \ourcolor \textbf{64.5} \\
\midrule
& \tocolor Target Only & \tocolor 76.1 & \tocolor  76.1 & \tocolor 76.1 & \tocolor 86.3 & \tocolor 86.3 & \tocolor 86.3 & \tocolor 72.8 & \tocolor 72.8 & \tocolor 72.8 & \tocolor 95.6 & \tocolor 95.6 & \tocolor 95.6 & \tocolor 82.7 \\

\bottomrule
\end{tabular}}
\end{center}
\vspace{-15pt}
\caption{VUDA results on \textit{Daily-DA}. Colored rows show our results, and the results in other rows are taken from \cite{li2023source} and \cite{unite}.
}
\vspace{-15pt}
\label{tab:uda_dailyDA}
\end{table*}

\begin{table}[t]
\begin{center}
\footnotesize
\resizebox{0.75\linewidth}{!}{
\begin{tabular}{clcc|c}
\toprule
& \multirow{2}{*}{\textbf{Method}} & \multicolumn{3}{c}{\textbf{Accuracy (Top-1\%)}} \\
& & H→U & U→H & \textbf{Avg.} \\
\midrule
&  \socolor Source Only & \socolor 93.3& \socolor 81.9 & \socolor 87.6 \\
\midrule
\scriptsize \parbox[t]{2mm}{\multirow{1}{*}{\rotatebox{90}{ZS}}}
& CLIP (ViT-B/16) \cite{radford2021learning} & 88.8 & 91.7 & 90.3 \\
\midrule
\scriptsize \parbox[t]{2mm}{\multirow{4}{*}{\rotatebox{90}{SFVUDA}}} & ATCoN \cite{atcon} & 85.3 & 79.7 & 82.5 \\
& EXTERN \cite{extern} & 91.9 & {88.9} & 90.4 \\
& STHC \cite{li2023source}& 92.1 & {90.9} & {91.5} \\
& DALL-V \cite{zara2023unreasonable} & 93.1 & {88.9} & {91.0} \\
\midrule
\scriptsize \parbox[t]{2mm}{\multirow{7}{*}{\rotatebox{90}{VUDA}}} & DANN \cite{dann} & 74.4 & 75.1 & 74.8 \\
& MK-MMD \cite{mkmdd} & 74.7 & 79.7 & 77.2 \\
& TA$^3$N \cite{chen2019temporal} & 78.1 & 84.8 & 81.5 \\
& CO$^2$A \cite{da2022dual} & 95.8 & 87.8 & 91.8 \\
& UDAVT \cite{dacostaUnsupervisedDomainAdaptation2022} & 96.8 & 92.3 & 94.6 \\
& UNITE \cite{unite} & 92.5 & \textbf{95.0} & 93.8 \\
& \ourcolor Ours w/o LMFT & \ourcolor 98.3 & \ourcolor 92.5 & \ourcolor 95.4 \\
& \ourcolor Ours& \ourcolor \textbf{98.6} & \ourcolor 94.2 & \ourcolor \textbf{96.4} \\
\midrule
& \tocolor Target Only & \tocolor 98.9 & \tocolor 97.2 & \tocolor 98.1 \\
\bottomrule
\end{tabular}}
\end{center}
\vspace{-15pt}
\caption{VUDA results on {\textit{UCF$\leftrightarrow$HMDB}\(_{full}\)}. 
Rows in colors are our experimental results, and others are from \cite{li2023source} and \cite{unite}.
}
\vspace{-2mm}
\label{tab:uda_ucf-hmdb}
\end{table}

\begin{table}[ht]
\begin{center}
\small
\resizebox{1\linewidth}{!}
{
\begin{tabular}{lccccccc|c}
\toprule
\multirow{2}{*}{\textbf{Method}} & \multicolumn{8}{c}{\textbf{Accuracy (Top-1\%)}} \\
 & KT→C1 & KT→C2 & KT→C3 & KT→C4 & KT→C5 & KT→C6 & KT→C7 & \textbf{Avg.} \\
\midrule
\socolor Source Only & \socolor 5.6& \socolor 8.7 & \socolor 100.0 & \socolor 65.9 & \socolor 89.5 & \socolor 100.0 & \socolor 100.0 & \socolor 67.1\\
\midrule
MM-SADA \cite{munro2020multi} & 11.0 & 0.0 & 100.0 & 72.9 & 68.4 & 100.0 & 93.2 & 62.6 \\
Zhang \textit{et al.} \cite{zhang2022audio} & 16.6 & 8.5 & 100.0 & 75.5 & 94.6 & 100.0 & 96.5 & 67.3 \\
UNITE \cite{unite} & 5.6 & 8.7 & 100.0 & \textbf{100.0} & \textbf{94.7} & 100.0 & \textbf{100.0} & 76.4 \\
\ourcolor Ours w/o LMFT & \ourcolor 77.8 & \ourcolor 69.6 & \ourcolor 100.0 &\ourcolor 75.6 &\ourcolor 79.0 &\ourcolor 100.0 &\ourcolor 100.0 &\ourcolor 84.8\\
\ourcolor Ours& \ourcolor \textbf{88.9} & \ourcolor \textbf{73.9} & \ourcolor 100.0 &\ourcolor 80.5 &\ourcolor 89.5 &\ourcolor 100.0 &\ourcolor 96.7 &\ourcolor \textbf{88.4}\\
\midrule
\tocolor Target Only & \tocolor 83.3 & \tocolor 65.2 & \tocolor 100.0 & \tocolor 78.1 & \tocolor 57.9 & \tocolor 88.9 & \tocolor 93.3 & \tocolor 81.1  \\
\bottomrule
\end{tabular}}
\end{center}
\vspace{-6mm}
\caption{VUDA results on {\textit{ActorShift}}. 
Colored rows indicate our experimental results. Uncolored baseline results are from \cite{zhang2022audio}, while UNITE \cite{unite} accuracy is from our run of their public code.}
\vspace{-6mm}
\label{tab:uda_actorshift}
\end{table}

\subsection{Datasets}
We use three VUDA benchmarks to evaluate our method:
\begin{itemize}
    \item \textbf{\textit{Daily-DA}} \cite{xu2023multi}: It contains 18,949 videos from four different domains: ARID (A) \cite{arid}, HMDB51 (H) \cite{hmdb51}, Moments-in-Time (M) \cite{mit} and Kinetics-600 (K) \cite{kinetics}. With 8 mutual classes, this benchmark represents human daily actions. Note that ARID was distinctly recorded under low illumination, making the task more challenging.
    \item \textbf{\textit{UCF-HMDB\textsubscript{\textit{full}}}} \cite{chen2019temporal}: It consists of 3,209 videos spanning 12 human common action classes from the HMDB51 (H) \cite{hmdb51} and UCF101 (U) \cite{ucf101} datasets.
    \item \textbf{\textit{ActorShift}}  \cite{zhang2022audio}: While benchmarks above are popular for evaluating video DA, they primarily focus on human-centric actions across domains. 
    To investigate adaptation beyond human-centered context, we utilize the \textbf{\textit{ActorShift}} dataset. This dataset presents a challenging scenario of adapting from human actions (Source: Kinetics-700 \cite{kinetics}) to the same actions performed by animals (Target: YouTube-collected videos), thereby creating a pronounced semantic gap. It comprises 1,505 videos across seven action categories: running (C1), drinking (C2), swimming (C3), eating (C4), sleeping (C5), opening a door (C6), and watching TV (C7). Critically, it also simulates a low-resource DA setting, requiring adaptation with only a limited number of unlabeled target samples.
\end{itemize}

\subsection{Implementation Details}
\label{sec:implement_details}
\noindent\textbf{Network architecture.}
We employ a ViT-B/16 architecture \cite{dosovitskiy2020image}, initialized with pretrained VideoMAE weights \cite{wang2023videomae}. 
To address the sequence length variability introduced by LMFT, we employ block-diagonal attention masking \cite{rlt}, which facilitates efficient batched processing by concatenating sequences without padding while strictly confining attention in ViT to per-video boundaries.
We conduct all experiments on NVIDIA L40 and RTX A6000 GPUs and leverage Flash Attention \cite{dao2022flashattention, dao2023flashattention} for computational efficiency.

\noindent\textbf{Setup.}
Following \cite{tong2022videomae}, we sample 16 frames per video with a tubelet size of $t_p=2$, resulting in a temporal resolution of $N_t=8$.
We resize all frames to $224 \times 224$ pixels. 
We train our model for 20 epochs using AdamW with a weight decay of 0.05 and a batch size of 32.
The initial policy parameters $\mu$ and $\log\sigma$ are set to 0.01 and -1.0, respectively.
The pseudo-labeling confidence threshold $\gamma_c$ and the target domain loss weight $\lambda_t$ are set to $0.8$ and $0.5$, respectively. In RL training,  $\lambda_{\mathcal{L}}$ is set to 10.
Following prior work in VUDA, we report Top-1\% accuracy on the validation set of the target domain.

\subsection{Comparison with the State-of-the-Art}
\noindent\textbf{Comparison Methods.}
We comprehensively evaluate the proposed framework against state-of-the-art methods on the above datasets. For \textit{Daily-DA}, we compare it with state-of-the-art VUDA methods, including DANN \cite{dann}, MK-MMD \cite{mkmdd}, TA$^3$N \cite{chen2019temporal}, and UNITE \cite{unite}. Additionally, we compare it with source-free VUDA (i.e., SFVUDA) approaches, including ATCoN \cite{atcon}, EXTERN \cite{extern}, STHC \cite{li2023source}, and DALL-V \cite{zara2023unreasonable}. We also report zero-shot accuracies from CLIP (ViT-B/16).
For \textit{UCF-HMDB\textsubscript{\textit{full}}}, we further include results from CO$^2$A \cite{da2022dual} and UDAVT \cite{dacostaUnsupervisedDomainAdaptation2022}.
For \textit{ActorShift}, we compare with MM-SADA \cite{munro2020multi}, Zhang \textit{et al.} \cite{zhang2022audio}, and UNITE \cite{unite}.
Moreover, we establish Source Only and Target Only baselines as the lower and upper bounds of our approach. For both, we fine-tune a ViT-B/16 backbone initialized with pretrained VideoMAE weights, on the source domain for Source Only and on the target domain for Target Only, then evaluate the resulting models on the target data.

\noindent\textbf{Main Results and Analysis.}
We present the results of our method on \textit{Daily-DA} in Table~\ref{tab:uda_dailyDA}, \textit{UCF-HMDB\textsubscript{\textit{full}}} in Table~\ref{tab:uda_ucf-hmdb}, and \textit{ActorShift} in Table~\ref{tab:uda_actorshift}.
To highlight the impact of LMFT, we also report results without LMFT (Ours w/o LMFT). In this setting, no tokens are dropped from source and target video frames; that is, all tokens are used in VUDA. We observe strong performance for Ours w/o LMFT, largely due to the high-quality pseudo labels generated by our confidence-based filtering step. With LMFT (Ours), we observe further substantial gains, and it is much more efficient due to token dropping (Sec. \ref{sec:ablations}). Across all three benchmarks, our method consistently outperforms state-of-the-art approaches by a significant margin: 5.3\% on \textit{Daily-DA}, 1.8\% on \textit{UCF-HMDB\textsubscript{\textit{full}}}, and 12\% on \textit{ActorShift}. On \textit{ActorShift}, our method even surpasses the Target Only baseline by 7.3\%. We attribute this superior performance to the benefits of DA, particularly in data-scarce target-domain scenarios. With limited target samples, the Target Only model tends to overfit; in contrast, our method effectively leverages both source and target domains to learn more robust and generalizable features, resulting in superior target domain performance.

\subsection{Efficiency Analysis and Ablations}
\label{sec:ablations}
We conduct a comprehensive efficiency analysis and several ablations on the M and H datasets from \textit{Daily-DA}.

\noindent\textbf{LMFT vs. other token reduction methods (training efficiency).}
To evaluate the training efficiency of LMFT for DA, we compare accuracy, training time, and the resulting speedup when substituting LMFT in our VUDA framework against five alternative methods:
i) Standard (\textit{ViT-B/16}): the default tokenization scheme in ViT, which corresponds to Ours w/o LMFT; 
ii) Random Dropping (\textit{Random}): a strategy that removes a fixed number of tokens per video, disregarding content;
iii) Token Merging (\textit{ToMe}) \cite{bolya2022token}: a popular token reduction technique in ViT that merges tokens with similar feature representations to reduce sequence length;
iv) \textit{PruMerge} \cite{prumerge}: a recent token reduction method which prunes tokens based on attention scores in ViT and merges them into the remaining ones; and 
v) \textit{DivPrune} \cite{divprune}: a state-of-the-art token reduction method that prunes the token sequence by promoting token diversity using cosine distance.
For a fair comparison, we apply a similar token-dropping ratio for all methods, consistent with LMFT. 
It is also worth noting that these reduction methods rely on manually tuned pruning ratios or thresholding rules for token reduction, whereas LMFT learns a threshold via RL to retain relevant tokens and discard the rest.
Table~\ref{tab:ab_training} shows that LMFT achieves the best VUDA performance-speed tradeoff.
\begin{table}[ht]
    \vspace{-3pt}
    \centering
    \footnotesize
    \setlength{\tabcolsep}{5.5pt}
    \begin{tabular}{lcccc}
        \toprule
        & Model & Accuracy ↑ & Train time ↓ & Speedup ↑ \\
        \midrule
        % ----- M→H -----
        \multirow{6}{*}{\scriptsize\rotatebox{90}{M→H}} 
        & \textcolor{gray}{ViT-B/16 \cite{dosovitskiy2020image}} & \textcolor{gray}{72.1} & \textcolor{gray}{3810s} & \textcolor{gray}{1.0$\times$} \\
        & Random & 71.7 & 3209s & 1.2$\times$ \\
        & ToMe \cite{bolya2022token} & 72.5 & 4117s & 0.9$\times$ \\
        & PruMerge \cite{prumerge} & 71.7 & 3630s & 1.1$\times$ \\
        & DivPrune \cite{divprune} & 72.9 & 16961s & 0.22$\times$ \\
        & \cellcolor{gray!20!white}\textbf{LMFT (Ours)} & \cellcolor{gray!20!white}\textbf{74.2} & \cellcolor{gray!20!white}\textbf{2784s} & \cellcolor{gray!20!white}\textbf{1.4}$\times$ \\
        \midrule
        % ----- H→M -----
        \multirow{6}{*}{\scriptsize\rotatebox{90}{H→M}} 
        & \textcolor{gray}{ViT-B/16 \cite{dosovitskiy2020image}} & \textcolor{gray}{57.5} & \textcolor{gray}{1211s} & \textcolor{gray}{1.0$\times$} \\
        & Random & 56.5 & 1016s & 1.2$\times$ \\
        & ToMe \cite{bolya2022token} & 59.0 & 1194s & 1.0$\times$ \\
        & PruMerge \cite{prumerge} & 57.9 & 930s & 1.3$\times$ \\
        & DivPrune \cite{divprune} & 56.5 & 4141s & 0.29$\times$ \\
        & \cellcolor{gray!20!white}\textbf{LMFT (Ours)} & \cellcolor{gray!20!white}\textbf{60.0} & \cellcolor{gray!20!white}\textbf{890s} & \cellcolor{gray!20!white}\textbf{1.4}$\times$ \\
        \bottomrule
    \end{tabular}
    \vspace{-5pt}
    \caption{Training efficiency of token reduction methods.
    }
    \vspace{-6pt}
    \label{tab:ab_training}
\end{table}

\noindent\textbf{LMFT vs. other token reduction methods (inference efficiency).}
We compare the inference-time efficiency of LMFT in terms of accuracy, throughput (Clips/s, where each clip consists of 16 frames), GFLOPs, and relative computational cost against standard tokenization (\textit{ViT-B/16}) in ViT, random token dropping (\textit{Random}), and three state-of-the-art token reduction techniques: \textit{ToMe} \cite{bolya2022token}, \textit{PruMerge} \cite{prumerge}, and \textit{DivPrune} \cite{divprune}. 
Table \ref{tab:ab_inference} shows that LMFT outperforms all other methods at inference, achieving superior adaptation while reducing computational cost.
\begin{table}[ht]
    \vspace{-5pt}
    \centering
    \footnotesize
    \setlength{\tabcolsep}{3.5pt}
    \begin{tabular}{lcccccc}
        \toprule
        & Model & Accuracy ↑ & Clips/s ↑ & GFLOPs ↓ & Cost ↓ \\
        \midrule
        % ----- M→H -----
        \multirow{6}{*}{\scriptsize\rotatebox{90}{M→H}} 
        & \textcolor{gray}{ViT-B/16 \cite{dosovitskiy2020image}} & \textcolor{gray}{72.1} & \textcolor{gray}{3.8} & \textcolor{gray}{266} & \textcolor{gray}{1.00$\times$} \\
        & Random & 71.7 & 3.6 & 216 & 0.81$\times$ \\
        & ToMe \cite{bolya2022token} & 72.5 & 1.8 & 242 & 0.91$\times$ \\
        & PruMerge \cite{prumerge} & 71.7 & 3.5 & 493 & 1.85$\times$ \\
        & DivPrune \cite{divprune} & 72.9 & 0.2 & 223 & 0.84$\times$ \\
        & \cellcolor{gray!20!white}\textbf{LMFT (Ours)} & 
          \cellcolor{gray!20!white}\textbf{74.2} & 
          \cellcolor{gray!20!white}\textbf{3.9} & 
          \cellcolor{gray!20!white}\textbf{217} & 
          \cellcolor{gray!20!white}\textbf{0.82}$\times$ \\
        \midrule
        % ----- H→M -----
        \multirow{6}{*}{\scriptsize\rotatebox{90}{H→M}} 
        & \textcolor{gray}{ViT-B/16 \cite{dosovitskiy2020image}} & \textcolor{gray}{57.5} & \textcolor{gray}{3.5} & \textcolor{gray}{266} & \textcolor{gray}{1.00$\times$} \\
        & Random & 56.5 & 3.7 & 218 & 0.82$\times$ \\
        & ToMe \cite{bolya2022token} & 59.0 & 2.2 & 243 & 0.91$\times$ \\
        & PruMerge \cite{prumerge} & 57.9 & 2.8 & 496 & 1.86$\times$ \\
        & DivPrune \cite{divprune} & 56.5 & 0.2 & 226 & 0.85$\times$ \\
        & \cellcolor{gray!20!white}\textbf{LMFT (Ours)} & 
          \cellcolor{gray!20!white}\textbf{60.0} & 
          \cellcolor{gray!20!white}\textbf{3.7} & 
          \cellcolor{gray!20!white}\textbf{218} & 
          \cellcolor{gray!20!white}\textbf{0.82}$\times$ \\
        \bottomrule
    \end{tabular}
    \vspace{-5pt}
    \caption{Inference efficiency of token reduction methods.}
    \vspace{-5pt}
    \label{tab:ab_inference}
\end{table}

\noindent\textbf{Efficiency comparison of VUDA methods.}
Our method not only achieves state-of-the-art performance but is also optimized for efficient VUDA. We compare it against top-performing VUDA methods, including UNITE \cite{unite} and DALL-V \cite{zara2023unreasonable}, and the standard tokenization baseline (\textit{ViT-B/16}). Note that DALL-V is a top source-free VUDA method. We compare in terms of accuracy, training time, and inference GFLOPs.
Table~\ref{tab:ab_da_methods} shows that our method achieves higher accuracy while reducing training time by approximately 10–20 times compared to the best VUDA method, UNITE. These results indicate that our method is both more effective and efficient. Note that since DALL-V and UNITE do not report efficiency analysis, we run their public code to obtain the results in the table.
\begin{table}[ht]
    \centering
    \footnotesize
    \setlength{\tabcolsep}{6pt}
    \begin{tabular}{lccccc}
        \toprule
        & Model & Accuracy ↑ & Train time ↓ & GFLOPs ↓\\
        \midrule
        % ----- M→H -----
        \multirow{4}{*}{\scriptsize\rotatebox{90}{M→H}} 
        & \textcolor{gray}{ViT-B/16 \cite{dosovitskiy2020image}} & \textcolor{gray}{72.1} & \textcolor{gray}{3810s} & \textcolor{gray}{266}  \\
        & DALL-V \cite{zara2023unreasonable} & 58.3 & 15526s & 291  \\
        & UNITE \cite{unite} & 71.7 & 27426s & 358  \\
        & \cellcolor{gray!20!white}\textbf{Ours} & 
        \cellcolor{gray!20!white}\textbf{74.2} & 
          \cellcolor{gray!20!white}\textbf{2784s} & 
          \cellcolor{gray!20!white}\textbf{217}  \\
        \midrule
        % ----- H→M -----
        \multirow{4}{*}{\scriptsize\rotatebox{90}{H→M}} 
        & \textcolor{gray}{ViT-B/16 \cite{dosovitskiy2020image}} & \textcolor{gray}{57.5} & \textcolor{gray}{1211s} & \textcolor{gray}{266} \\
        & DALL-V \cite{zara2023unreasonable} & 46.8 & 18368s & 291  \\
        & UNITE \cite{unite} & 49.0 & 22070s & 358  \\
        & \cellcolor{gray!20!white}\textbf{Ours} & 
          \cellcolor{gray!20!white}\textbf{60.0} & 
          \cellcolor{gray!20!white}\textbf{890s} & 
          \cellcolor{gray!20!white}\textbf{218} \\
        \bottomrule
    \end{tabular}
    \vspace{-5pt}
    \caption{Efficiency comparison of VUDA methods.}
    \vspace{-12pt}
    \label{tab:ab_da_methods}
\end{table}

\begin{figure*}[t]
    \centering
    \includegraphics[width=\textwidth, height=4.8cm]{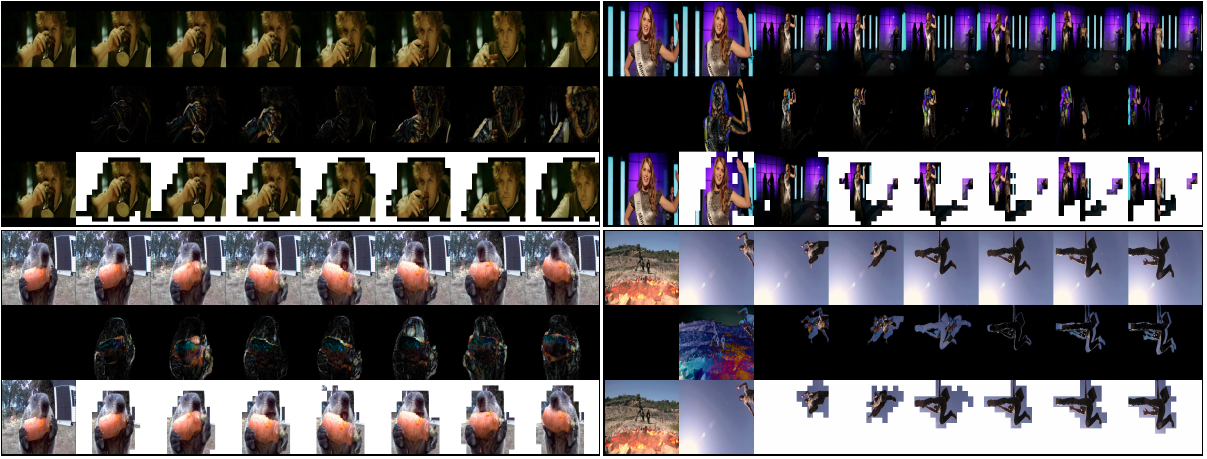}
    \vspace{-20pt}
    \caption{
    Visualization of LMFT on four videos (two left, two right).
    Each video has three rows: original frames, motion differences, and LMFT processed frames. LMFT selects action-relevant patches, dropping static or low-motion background patches for effective DA.
    }
    \vspace{-12pt}
    \label{figure:ab_visualization}
\end{figure*}

\noindent\textbf{Confidence-based filtering.}
We perform confidence-based filtering and use the high-confidence target domain pseudo-labels from the zero-shot CLIP model for training.
To assess the impact of pseudo-label quality, we vary the confidence threshold $\gamma_c \in \{0.0, 0.2, 0.4, 0.6, 0.8, 1.0\}$
and evaluate performance. When $\gamma_c$ is 1.0, no pseudo-labels are selected for the target, resulting in the Source Only setting.
Table~\ref{tab:ab_confidence_thres} shows that $\gamma_c=0.8$ offers the optimal fidelity-accuracy trade-off; thus, we fix $\gamma_c=0.8$ in our experiments.
\begin{table}[ht]
\vspace{-5pt}
\begin{center}
\footnotesize
\setlength{\tabcolsep}{12pt}
\resizebox{0.85\linewidth}{!}{
\begin{tabular}{ccc} 
\toprule
\multirow{2}{*}{\textbf{Confidence Threshold $\gamma_c$}} & \multicolumn{2}{c}{\textbf{Accuracy (\%)}} \\
 & M$\rightarrow$H & H$\rightarrow$M \\
\midrule
0.0 & 68.8 & 54.3 \\ 
0.2 & 67.5 & 57.0 \\
0.4 & 68.8 & 52.7 \\
0.6 & 72.5 & 57.7 \\
\cellcolor{gray!20!white}\textbf{0.8}& \cellcolor{gray!20!white}\textbf{74.2} & \cellcolor{gray!20!white}\textbf{60.0} \\
1.0 & 68.8 & 47.5 \\
\bottomrule
\end{tabular}}
\end{center}
\vspace{-15pt}
\caption{Impact of threshold $\gamma_c$ in confidence-based filtering.}
\vspace{-5pt}
\label{tab:ab_confidence_thres}
\end{table}

\noindent\textbf{Impact of different temporal resolution.}
To evaluate robustness under reduced temporal evidence, we vary $N_t$ (see Sec. \ref{sec:preliminaries} and Sec. \ref{sec:implement_details}), with $N_t \in \{2, 4, 6, 8\}$. 
We benchmark our method against three schemes: standard tokenization (\textit{ViT-B/16}), random dropping (\textit{Random}), and the state-of-the-art method, UNITE~\cite{unite}.
Table~\ref{tab:ab_fewer_frames} shows that our method consistently outperforms all baselines across all temporal budgets, demonstrating strong robustness even when the available temporal context is reduced. This result suggests that LMFT effectively captures action-relevant regions from limited frames, enabling efficient and reliable adaptation under temporally constrained conditions. Note that we run the public code of UNITE to obtain the results in the table.
\begin{table}[ht]
    \vspace{-5pt}
    \centering
    \footnotesize
    \setlength{\tabcolsep}{6pt}
    \begin{tabular}{lcccccc}
        \toprule
        & Model & $N_t=2$ & $N_t=4$ & $N_t=6$ & $N_t=8$\\
        \midrule
        % ----- M→H -----
        \multirow{4}{*}{\scriptsize\rotatebox{90}{M→H}} 
        & \textcolor{gray}{ViT-B/16 \cite{dosovitskiy2020image}} & \textcolor{gray}{64.2} & \textcolor{gray}{69.2} & \textcolor{gray}{70.4} & \textcolor{gray}{72.1} \\
        & Random & 63.3 & 67.5 & 70.4 & 71.7 \\
        & UNITE \cite{unite} & 63.3 & 70.4 & 68.8 & 71.7  \\
        & \cellcolor{gray!20!white}\textbf{Ours} & 
        \cellcolor{gray!20!white}\textbf{65.4} & 
          \cellcolor{gray!20!white}\textbf{71.3} & 
          \cellcolor{gray!20!white}\textbf{73.3} & 
          \cellcolor{gray!20!white}\textbf{74.2} \\
        \midrule
        % ----- H→M -----
        \multirow{4}{*}{\scriptsize\rotatebox{90}{H→M}} 
        & \textcolor{gray}{ViT-B/16 \cite{dosovitskiy2020image}} & \textcolor{gray}{48.0} & \textcolor{gray}{50.0} & \textcolor{gray}{53.5} & \textcolor{gray}{57.5} \\
        & Random & 46.7 & 52.0 & 52.5 & 56.5 \\
        & UNITE \cite{unite} & 41.3 & 43.3 & 43.3 & 49.0 \\
        & \cellcolor{gray!20!white}\textbf{Ours} & 
          \cellcolor{gray!20!white}\textbf{49.8} & 
          \cellcolor{gray!20!white}\textbf{52.3} & 
          \cellcolor{gray!20!white}\textbf{54.0} & 
          \cellcolor{gray!20!white}\textbf{60.0} \\
        \bottomrule
    \end{tabular}
    \vspace{-5pt}
    \caption{Impact of different temporal resolutions on VUDA.}
    \vspace{-8pt}
    \label{tab:ab_fewer_frames}
\end{table}

\noindent\textbf{Effect of the Reward Coefficient.}
We evaluate the sensitivity of the reward formulation (Eq.~\ref{eq:reward}) to the coefficient $\lambda_{\mathcal{L}}$ across values $\{0.1, 1.0, 10.0, 100.0\}$. Results in Table~\ref{tab:ab_lamda_l} show that the overall performance remains stable, indicating the policy optimization is not overly dependent on a precise choice of this hyperparameter. Given that $\lambda_{\mathcal{L}} = 10.0$ yields the most balanced results across both adaptation settings, we therefore adopt $\lambda_{\mathcal{L}} = 10.0$ for all experiments.
\begin{table}[ht]
\vspace{-5pt}
\begin{center}
\setlength{\tabcolsep}{15pt}
\resizebox{0.9\linewidth}{!}{
\begin{tabular}{ccc} 
\toprule
\multirow{2}{*}{\textbf{Reward hyperparameter $\lambda_{\mathcal{L}}$}} & \multicolumn{2}{c}{\textbf{Accuracy (\%)}} \\
 & M$\rightarrow$H & H$\rightarrow$M \\
\midrule
0.1 & 73.3 & 56.7\\
1.0 & 73.8 & 57.0\\
\cellcolor{gray!20!white}\textbf{10.0 }& \cellcolor{gray!20!white}\textbf{74.2} & \cellcolor{gray!20!white}\textbf{60.0} \\
100.0 & 72.8 & 58.5 \\
\bottomrule
\end{tabular}}
\end{center}
\vspace{-15pt}
\caption{Effect of varying $\lambda_{\mathcal{L}}$ in Eq.~\ref{eq:reward}.}
\vspace{-8pt}
\label{tab:ab_lamda_l}
\end{table}

\noindent\textbf{Comparison with Gumbel-Softmax.}
Instead of using RL to learn the motion-focused threshold ($\tau$) in Eq.~\ref{eq:selection_mask}, we could use Gumbel-Softmax \cite{maddison2016concrete, 2025bypass} since it provides a differentiable approximation for discrete choices via continuous relaxation. 
We compare LMFT with Gumbel-Softmax in terms of minimum (Min) and maximum (Max) memory consumption during training, training time, and accuracy. 
Table~\ref{tab:gumbelsoft} shows that Gumbel-softmax underperforms and is less computationally efficient than LMFT, indicating that RL provides a more effective mechanism for learning $\tau$.
\begin{table}[ht]
\vspace{-5pt}
\begin{center}
% \small
\footnotesize
\resizebox{\linewidth}{!}{
\begin{tabular}{lccccc}
\toprule
& \multirow{2}{*}{\textbf{Method}} & \multicolumn{2}{c}{\textbf{Memory (GiB) ↓}} & \multirow{2}{*}{\textbf{Time ↓}} & \multirow{2}{*}{\textbf{Accuracy ↑}} \\
 & & Min & Max &  &  \\
\midrule
\multirow{2}{*}{\rotatebox{90}{M→H}} & Gumbel-Softmax & 1.31 & 34.58 & 3275 & 73.3 \\
 & \cellcolor{gray!20!white}\textbf{LMFT(Ours)} & \cellcolor{gray!20!white}\textbf{1.31} & \cellcolor{gray!20!white}\textbf{30.03} & \cellcolor{gray!20!white}\textbf{2784} & \cellcolor{gray!20!white}\textbf{74.2} \\
\midrule
\multirow{2}{*}{\rotatebox{90}{H→M}} & Gumbel-Softmax & 1.31 & 34.58 & 1130 & 58.8 \\
 & \cellcolor{gray!20!white}\textbf{LMFT(Ours)} & \cellcolor{gray!20!white}\textbf{1.31} & \cellcolor{gray!20!white}\textbf{28.96} & \cellcolor{gray!20!white}\textbf{890} & \cellcolor{gray!20!white}\textbf{60.0} \\
\bottomrule
\end{tabular}}
\end{center}
\vspace{-15pt}
\caption{Gumbel-Softmax vs. RL in LMFT.}
\vspace{-8pt}
\label{tab:gumbelsoft}
\end{table}

\noindent\textbf{Qualitative Analysis.}
In Fig.~\ref{figure:ab_visualization}, we qualitatively show the effectiveness of LMFT. In the first two videos (left), LMFT successfully identifies motion-rich, action-relevant patches while filtering out static, domain-specific background regions. In the latter two videos (right), even under substantial viewpoint variations, LMFT consistently selects action-centric regions. These visualizations highlight LMFT’s capability to select motion-rich, action-relevant regions and drop redundant, scene-specific content, thereby mitigating domain shift and improving VUDA for action recognition.

\section{Conclusion}
We introduce Learnable Motion-Focused Tokenization (LMFT) for effective and efficient VUDA. By learning to discard low-motion, redundant tokens while retaining motion-rich, action-relevant ones, LMFT facilitates action-centric video representations that better align across domains. Extensive experiments on three VUDA benchmarks demonstrate that our method establishes a new state of the art. Moreover, we provide the first comprehensive analysis of computational efficiency in VUDA, showing that LMFT significantly reduces computational cost for VUDA. These results underscore the importance of informative token selection and suggest that efficient tokenization strategies offer a promising path toward effective and computationally efficient video domain adaptation.

\noindent{\bf Acknowledgements:} 
We acknowledge the support of the University of Saskatchewan and the Natural Sciences and Engineering Research Council of Canada (NSERC).

{   
    \small
    \bibliographystyle{ieeenat_fullname}
    \bibliography{main}
}

\end{document}